# *Physarum* boats: If plasmodium sailed it would never leave a port

Andrew Adamatzky

University of the West of England, Bristol BS16 1QY, United Kingdom
andrew.adamatzky@uwe.ac.uk

October 25, 2018

## Abstract

Plasmodium of *Physarum polycephalum* is a single huge (visible by naked eye) cell with myriad of nuclei. The plasmodium is a promising substrate for non-classical, nature-inspired, computing devices. It is capable for approximation of shortest path, computation of planar proximity graphs and plane tessellations, primitive memory and decision-making. The unique properties of the plasmodium make it an ideal candidate for a role of amorphous biological robots with massive parallel information processing and distributed inputs and outputs. We show that when adhered to light-weight object resting on a water surface the plasmodium can propel the object by oscillating its protoplasmic pseudopodia. In experimental laboratory conditions and computational experiments we study phenomenology of the plasmodium-floater system, and possible mechanisms of controlling motion of objects propelled by on board plasmodium.

*K*eywords: *Physarum polycephalum*, motility, biological robots

## 1 Introduction

A plasmodium, or vegetative state, of *Physarum polycephalum*, also known as a true or multi-headed slime mold, is a single cell with many diploid nuclei; it behaves like a giant amoeba. In 2000 Nakagaki *et al* [17, 18, 19] shown that topology of the plasmodium's protoplasmic network optimizes the plasmodium's harvesting on distributed sources of nutrients and makes more efficient flow and transport of intra-cellular components. Thus the field of Physarum computing emerged [20]. The problems solved by the plasmodium include approximation of shortest path in a maze [18, 19], computation of proximity graphs [8], Delaunay triangulation [21], construction of logical gates [22], robot control [23], implementation of storage-modification machines [7], and approximation of Voronoi diagram [21].

The plasmodium can be seen as a network of biochemical oscillators [14, 16], where waves of excitation or contraction originate in several sources, travel along the plasmodium, and interact one with another in collisions. The oscillatory cytoplasm of the plasmodium is a spatially extended nonlinear excitable media. Previously we shown that the plasmodium of *Physarum* is a biological analogue of chemical reaction-diffusion system encapsulated in an elastic membrane [6]. Such encapsulation allows the plasmodium to act as a massive-parallel reaction-diffusion computer [4] with parallel inputs and outputs. This ensures the plasmodium explores environment in a distributed manner and responds to changes in surrounding conditions as an amorphous decentralized but yet internally coordinated entity. Thus we can envisage that the plasmodium of *Physarum polycephalum* is a promising biological substrate for designing amorphous robots with embedded reaction-diffusion intelligence. An engineering background for such amoeboid machines was developed by Yokoi *et al* [24, 25], however no viable non-silicon implementations are known so far.

There were just two attempts to integrate spatially extended non-linear chemical or biological systems with silicon hardware robots. In 2003 a wheeled robot controlled by on board excitable chemical system, the Belousov-Zhabotinsky medium, was successfully tested in experimental arena [3]. The on board chemical reaction was stimulated by silver wire and motion vector towards source of stimulation was exctracted by robot from topology of the excitation waves in the Belousov-Zhabotinsky medium. In 2006 Tsuda *et al* [23] designed and successfully tested in real-world conditions a *Physarum* contoller for a legged robot. The controller's functioning was based on the fact that light inhibits oscillations in illuminated parts of the plasmodium. Thus a direction towards light source can be calculated from phase differences between shaded and illuminated parts of the plasmodium controller. In both instances non-linear media controllers were coupled with conventional hardware and relied upon additional silicon devices to convert spatio-temporal dynamics of excitation and oscillation to the robots motion.

The only known up to date result of self-propelled excitable-medium devices can be attributed to Kitahata [13] who in 2005 experimentally shown that a droplet of Belousov-Zhabotinsky medium is capable for translation motion due to changes in interfacial tension caused by convention forces inside the droplet. The convention forces are induced by excitation waves, propagating inside the droplet [12].

Recently we experimentally demonstrated [10] that the plasmodium can act as a distributed manipulator when placed on a water surface with light-weight objects distributed around. We shown that the plasmodium senses data-objects, calculates shortest path between the objects, pushes and pulls light-weight objects placed on a water surface. We also found that motility of the plasmodium placed *directly* on the water surface is restricted [10]. The plasmodium does propagate pseudopodia and forms protoplasmic trees, however it does not travel as a localized entity. To achieve true mobility of a plasmodium we decided to attach it to a floater in a hope that due to mechanical oscillations of propagating pseudopodia the plasmodium will be capable for applying enough propulsive force to the floater to propel the floater on the water surface. In laboratory experiments we proved feasibility of the approach. In present paper we report our findings on motion patterns observed in experiments and simulations.

The paper is structured as follows. Experimental techniques and basics of simulating a plasmodium-floater system in mobile cellular automaton lattices are presented in Sect. 2. In Sect. 3 we discuss modes of movement of plasmodium-floater systems, observed in laboratory experiments. Results of computer simulations of the mobile excitable lattices, that mimic the plasmodium behavior, are shown in Sect. 4. We overview the results and discuss our findings in a context of unconventional robots and robotic controllers in Sect. 5.

## 2 Methods

Plasmodium of *Physarum polycephalum* was cultivated on a wet paper towels in dark ventilated containers. Oat flakes were supplied as a substrate for bacteria on which the plasmodium feeds. Experiments were undertaken in Petri dishes with base diameter 35 mm, filled by 1/3-1/5 with distilled water. Variously sized (but not exceeding 5-7 mm in longest dimension) pieces of plastic and foam were used as floaters residing on the water surface. Oat flakes occupied by the plasmodium were placed on top of the floaters. Behavior of the plasmodium-floater systems was recorded using QX-5 digital microscope, magnification ×10. Videos of experiments are available at [11].

The Petri dish with plasmodium was illuminated by LED (minimal light setting for the camera) by white light which acted as a stimulus for the plasmodium phototaxis. The illuminating LED was positioned 2-4 cm above North-East part of the experimental container. The microscope was placed in the dark box to keep illumination gradient in the dish safe from disturbances.

Plasmodium-floater systems were simulated by mobile two-dimensional cellular automata, with eight cell neighbourhood, which produce force fields [2]. We employed the model of retained excitation [5]. Every cell of the automaton takes three states — resting, excited and refractory, and updates its state depending on states of its eight closest neighbors. A resting cell becomes excited if number of its excited neighbors lies in the interval $[\theta_1, \theta_2]$, $1 \leq \theta_1, \theta_2 \leq 8$. An excited cell remains excited if number of excited neighbors lies in the interval $[\delta_1, \delta_2]$, $1 \leq \delta_1, \delta_2 \leq 8$, otherwise the cell takes refractory state. A cell in refractory state becomes resting in the next time step. This is a model of 'retained excitation' [5] used to imitate *Physarum* foraging behavior [6] because this cellular automaton exhibits 'amoeba-shaped' patterns of excitation. We denote the local transitions functions as $R(\theta_1, \theta_2, \delta_1, \delta_2)$.

The reaction to light stimulus was implemented as follows. Cells of lattice edges most distant from the light were excited with probability 0.15. The illumination dependent excitation of edge cells corresponds to stimulus-dependent changes in plasmodium oscillation frequencies [15, 16]. Namely, the parts of the plasmodium closest to positive stimulus (e.g. sources of nutrients) periodically contract with higher frequencies than parts closest to negative stimulus (e.g.

high illumination).

We derived motion from excitation dynamics by supplying every cell of the lattice with a virtual local force vector. The local vectors are updated at every step of the simulation. A vector in each cell becomes oriented towards less excited part of the cell's neighborhood. An integral vector, which determines lattice rotation and translation at each step of simulation time, is calculated as a sum of local vectors over all cells. The approach is proved to be successful in simulation of mobile excitable lattices, see details in [2].

## 3   Experimental results

In experiments, we observed five types of movements executed by plasmodium-floater system: random wandering, quick sliding, pushing, and directed propelling.

The random wandering is caused by sudden movements, and associated vibrations, of plasmodium's protoplasm. Movement of protoplasm is caused by peristaltic contraction waves generated by disordered network of biochemical oscillators. The higher the frequency of contractions in a particular domain of the plasmodium, the more protoplasm is attracted into the domain [15, 16]. Relocations of protoplasmic mass change mass center of the plasmodium-floater system, thus causing the floater to wander. The floater influenced by the protoplasmic vibration may exhibit a random motion (Fig. 1a–e), which usually persists till some part of the plasmodium propagates beyond edges of the floater (after that another types of motion take place).

The quick sliding motion occurs when substantial amount of protoplasm, e.g. a propagating pseudopodium, relocates to one edge of the floater and this pseudopodium penetrates water surface. The floater then become partly submerged. This changes the way water surface forces act on the floater, and sets the floater in a motion. The floater usually slides till collide with an obstacle or a wall of the container.

The random wandering and sliding are rather types of motion with non controllable and unpredictable trajectories. These motions unlikely to be used in future architectures of biological robots so we not discuss them further.

The pushing motion can be observed in situations when plasmodium propagates from the floater onto the water surface and develops a tree of protoplasmic tubes. At some stage of the development tips of the growing protoplasmic tree reach walls of the water container and adhere to the walls. An example is shown in Fig. 2: three-four hours after being placed on a floater the plasmodium starts propagation (Fig. 2bc) and a protoplasmic tree emerges (Fig. 2f). Already at this stage the floater comes into motion (Fig. 2fg), traveling in the direction opposite to propagation of the protoplasmic tree. The motion becomes pronounced when a part of the tree attaches itself to a wall of the container (Fig. 2h), a force of protoplasm pumped into the tubes causes the floater to noticeably move outwards the growing tubes (Fig. 2i).

A position of the floater can be 'fine-tuned' by several protoplasmic trees,

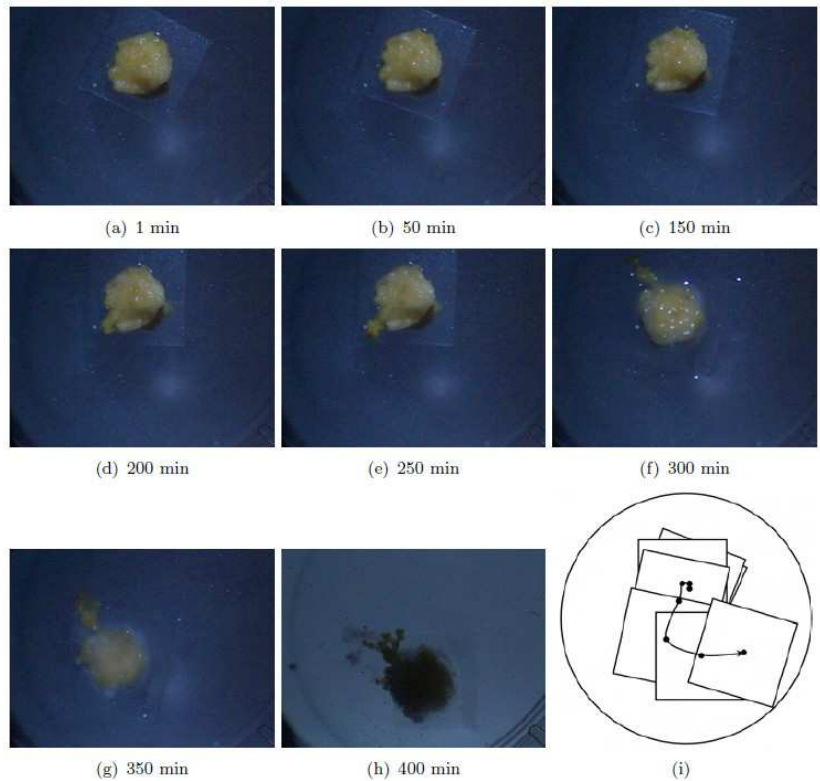

Figure 1: Transition between types of motion of physarum-floater system. Random wandering, or vibration-based, movement (a)–(e), $90^o$ clockwise rotation (e)–(f), and directed propelling (f)–(h). Positions of the floater, at time steps corresponding to snapshots (a)–(h) and trajectory the floater's centre are shown in (i). See video at [11].

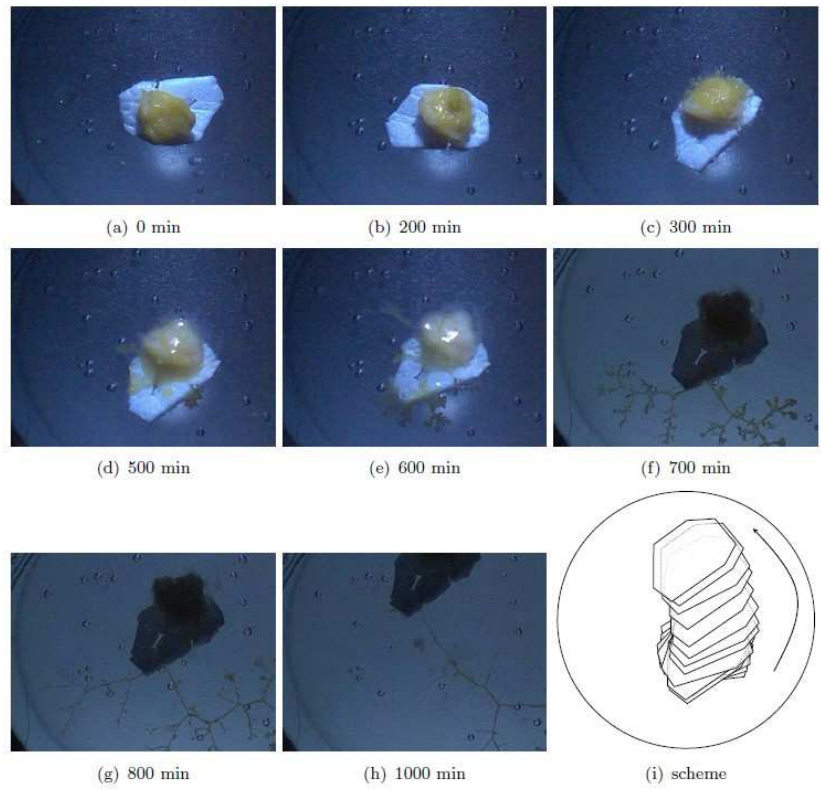

Figure 2: Pushing the floater by growing protoplasmic tubes. (a)–(h) Photographs of experimental container. (i) Time-lapsed contours of the floater. See video at [11].

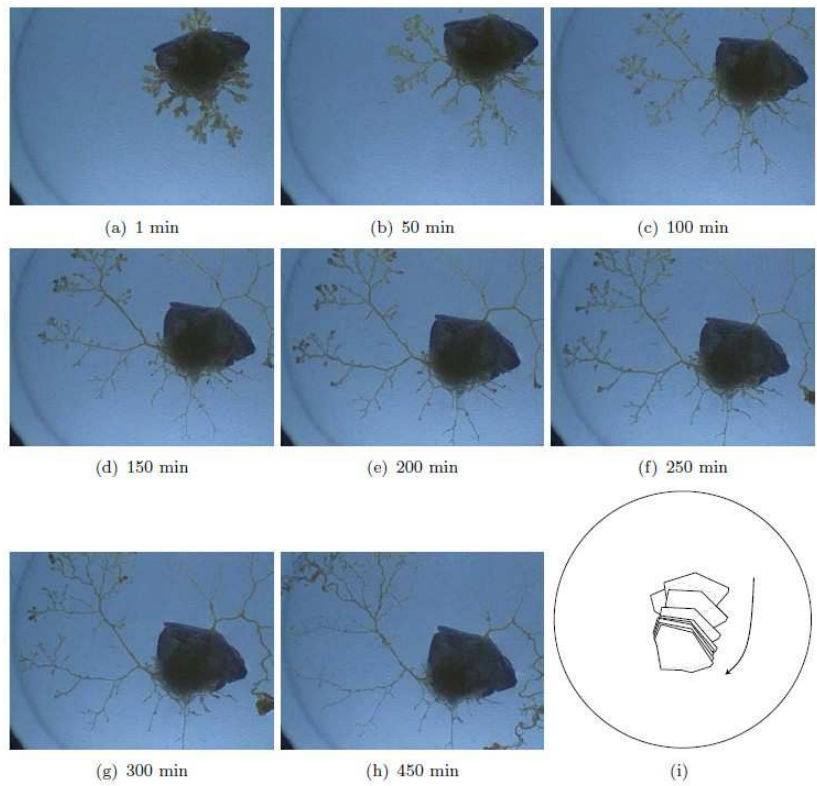

Figure 3: Collective movement of the floater by several protoplasmic trees. (a)–(h) Photographs of the experimental container. (i) Time-lapse contours of the floater. See video at [11].

which attached themselves to different sides of experimental container (Fig. 3). In experiment shown in Fig. 3 protoplasmic trees grow along North-West to South-East arc and to the direction North-East (Fig. 3bc). The sub-tree growing in the North-East direction diminishes with time, because it happen to be in the most illuminated area of the container, the only branch heading South survives. The protoplasmic tree becomes stronger in the North-West direction (Fig. 3bc)g–h). The dynamical reallocations of the trees cause the floater to rotate and then gradually move South-West (Fig. 3i).

The last and the most interesting type of motion observed is direct propelling of floater by on board plasmodium. This type of motion occurs when pseudopodia propagate beyond the edge of the floater but then partly sank. The pseudopodia oscillate, their oscillations are clearly visible on the video recordings in [11]. These oscillations cause the floater to move in the direction opposite to submersed pseudopodia (in contrast to sliding motion, the floater does not submerse at all during the propulsive motion). The examples of direct propelling are shown in Figs. 1 and 4.

In experiment shown in Fig. 1 plasmodium propagates a pseudopodium in South-West direction (Fig. 1de). The pseudopodium does not stay on the water surface but partly sinks. The initial wetting of the pseudopodium causes the floater to rotate clockwise with the protruding pseudopodium facing now North-West (Fig. 1de). The immersed pseudopodia oscillate and propel the floater in the South-Easterly direction (Fig. 1g–i).

An example of collective steering of the floater by several pseudopodia is shown in Fig. 4. After initial adaptation period (Fig. 4ab), associated with vibrational random motion, the plasmodium propagates pseudopodia in Westward and South-Easterly directions (Fig. 4de). At the beginning only South-Eastern pseudopodia contribute to propelling of the floater, with Western pseudopodia are merely balancing action of their South-Eastern counterparts. Thus the floater travels towards North (Fig. 4fg). Later Westerly pseudopodia increase frequency of their oscillations (see [11]) and the floater turn more towards North-East (Fig. 4hi).

## 4   Results of simulations

**Proposition 1.** *Given large enough container and a stationary source of light a floater with plasmodium on board will move in irregular cycles around the source.*

By assuming the container is large enough we secured the situation when it is impossible for pseudopodia, or protoplasmic trees, to reach walls of the container and push the floater. The wandering motion only occurs at the very beginning of plasmodium's development on the floater. Thus we are left with propelling motion.

Let the floater be stationary. Due to negative phototaxis, plasmodium's pseudopodia propagate towards the less illuminated parts of the container.

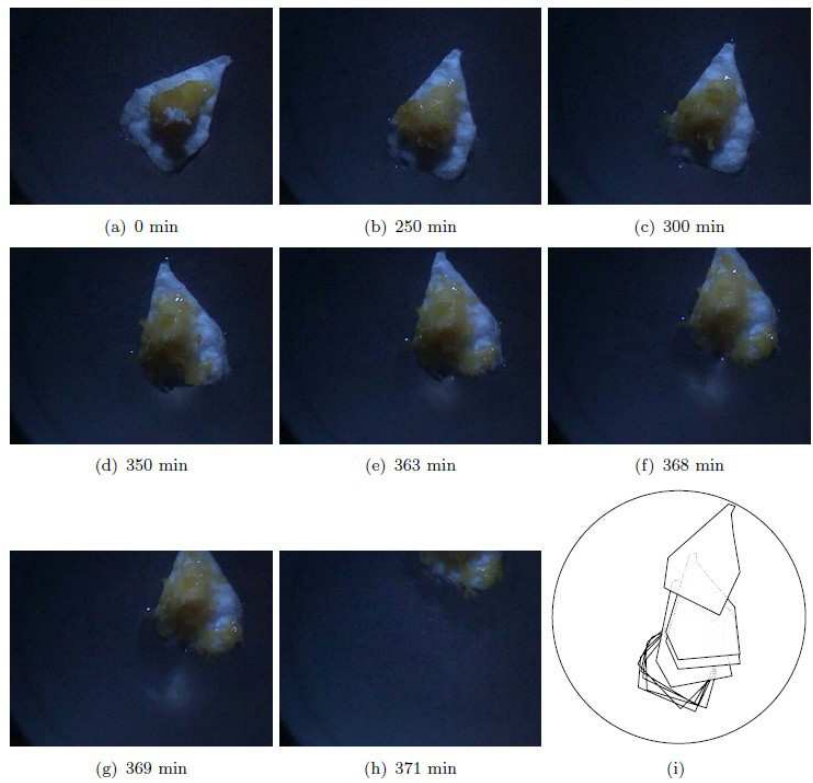

Figure 4: Direct propelling of floater by oscillating pseudopodia. (a)–(i) Photographs of the experimental container. (j) Time lapses contours of the floater. See video at [11].

When the pseudopodia spawn from the floater to the water surface, their oscillatory contractions propel the floater towards source of illumination. The pseudopodia continue their action till the floater pass the highly illuminated part. Then the pseudopodia happens to be on the most illuminated side of the floater. They retract and new pseudopodia are formed on the less illuminated part of the floater. They propel the floater towards the source of illumination again. Exact positions of growing pseudopodia depends on distribution of biochemical sources of oscillations and interactions between traveling waves of contraction. New pseudopodia may not emerge at the same place as old pseudopodia. Thus each trajectory of the floater towards the source of illumination will be different from previous ones.

To verify the proposition we simulated plasmodium-floater system by mobile cellular automata with various cell-state transition functions. An illustrative series of snapshots is shown in Fig. 5. The automaton lattice was placed at some distance from the source of light (Fig. 5a). Edges of the lattice oriented towards less illuminated area of experimental container became excited (Fig. 5b). The excitation propels the lattice towards the source of light (Fig. 5e–f). When lattice passes the site with highest illumination, another edge of the lattice excites and propels the lattice back to the domain of high illumination (Fig. 5gh). The sequence of excitations continues and the lattice travels along irregular cyclic trajectories around the source of illumination (Fig. 5i–k).

Exact pattern of lattice cycling around the source of illumination is determined by particulars of the local excitation dynamics. For example, trajectories of a lattice with threshold excitation (Fig. 6a) are very compactly arranged around the source: as soon as the lattice approaches the source of illumination it stays nearby, mostly turning back and force. When we impose upper boundary of the excitation, e.g. when a resting cell is excited only if one or two of its neighbors are excited, the lattice trajectory loosens (Fig. 6b).

Let us discuss now few examples of lattices governed by functions $R(\theta_1\theta_2\delta_1\delta_2)$, where a resting cell excites if number $\sigma$ of excited neighbors lies in the interval $[\theta_1, \theta_2]$ and excited cell can stay excited if $\sigma \in [\delta_1, \delta_2]$ (Fig. 6c–f). We focus on excitation interval $[2, 2]$ because this type of local transition is typical for media with traveling localized excitations [1], which does closely relate to propagation activities of *Physarum polycephalum* [9]. We observed that lattices with very narrow intervals of excitation and retained excitation demonstrate a combination of compact and lose trajectories (Fig. 6c). Widening interval of retained excitation disperses lattice trajectories in space (Fig. 6de), thus emulating delayed responses to changes in sensorial background. When lower, $\delta_1$, and upper, $\delta_2$, boundaries of retained excitation increase to 4 and 6, respectively, a lattice starts to exhibit a quasi-ordered behavior: combination of long runs away from the source and tidy loops of rotations (Fig. 6f).

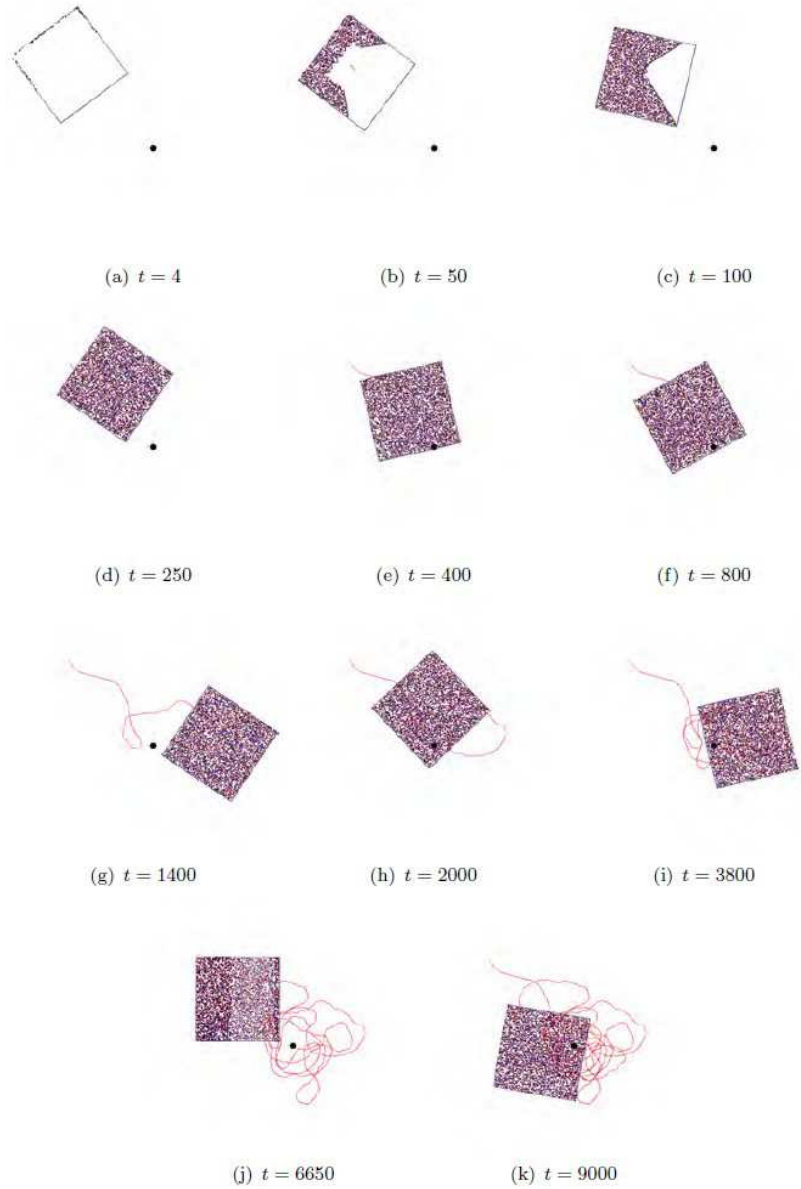

Figure 5: Snapshots of cellular automaton model (two-dimensional lattice of $200 \times 200$ cells) of plasmodium-floater system. The local excitation dynamics is controlled by function $R(2201)$. Source of illumination is shown by solid black disc. Each snapshot is supplied with trajectory of the center of the floater from the beginning of simulation. See exemplar videos at [11].

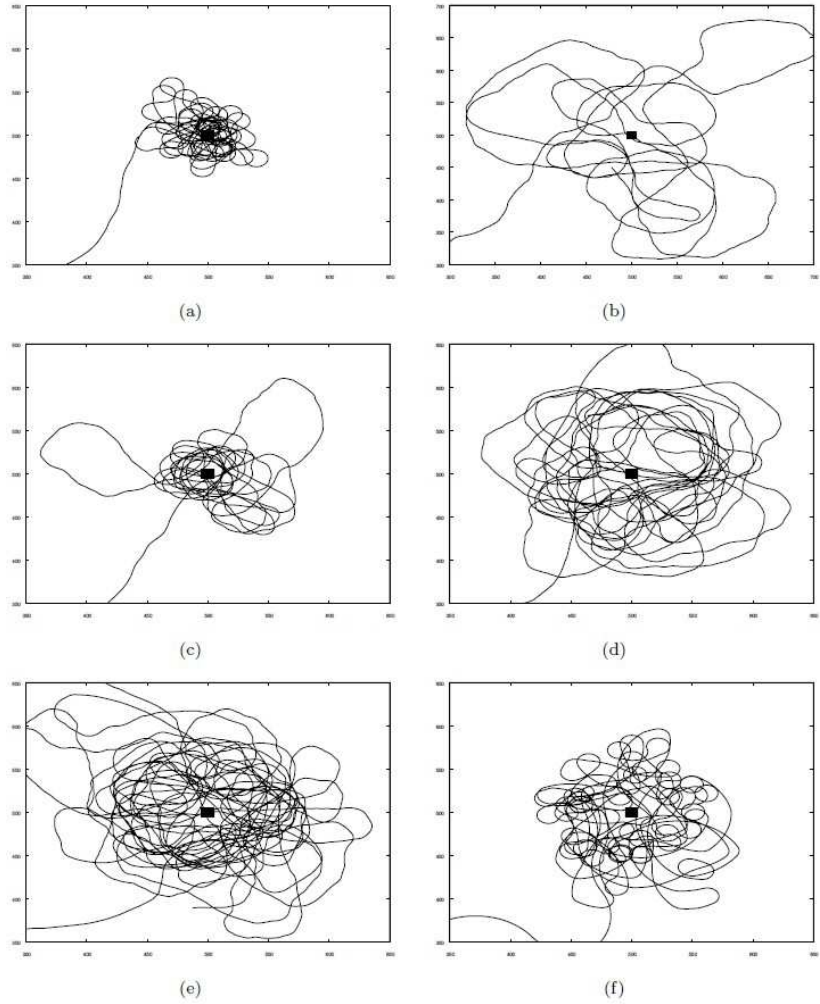

Figure 6: Trajectories of the center of cellular-automaton lattice traveling around a source of illumination. The lattice's behavior is governed by local transition functions as follows: (a) $R(1899)$, threshold excitation, a resting cell excites if there is at least one excited neighbor, values 9 for $\delta_1$ and $\delta_2$ mean that the cell never stays excited longer then one step of a time, (b) $R(1299)$, a resting cell excites if it has one or two excited neighbors. (c) $R(2222)$, (d) $R(2201)$, (e) $R(2211)$, (f) $R(2246)$,

## 5 Summary

In laboratory experiments and computer simulations we demonstrated that plasmodium of *Physarum polycephalum* can act as an 'engine' or a propelling agent for light-weight floating objects. We found that most typical type of movement generated by the plasmodium is a propulsive forward motion. This motion is caused by pseudopodia protruding beyond the the floater and oscillating. In small container, when growing tree of plasmodium's protoplasmic tubes can reach sides of the container, the plasmodium pushes floater by increasing length of the tubes connecting the float and the container's sides.

A plasmodium shows negative phototaxis. It tries to evade regions illuminated by non-yellow light and grows towards more shaded areas. When the plasmodium attached to a floating object, plasmodium–floater system exhibits positive phototaxis. Due to growth, and associated oscillations, of pseudopodia on the less illuminated side of a floater, the floater moves towards light. We mimicked this phenomenon in cellular automaton models of mobile light-sensitive lattices. We found that in ideal conditions the "plasmodium–floater" system will wander around the site of highest illumination, often following quasi-chaotic trajectories due to many sources of excitation competing with one another in the protoplasm.

The phenomena and primitive constructs of the plasmodium-floaters will be employed in future designs of amorphous decentralized robots operating on water-surface. The robots will be capable to search for objects, travel towards the objects' location, implement manipulation and sorting. The robots will be controlled by gradient fields of illumination and chemo-attractants.

## Acknowledgement

The works is supported by the Leverhulme Trust Research Grant F/00577/I.